\documentclass[letterpaper, 10 pt, conference]{ieeeconf}

\usepackage{amsmath}
\usepackage{algorithm}
\usepackage[noend]{algpseudocode}
\usepackage{stfloats} 
\IEEEoverridecommandlockouts                              

\usepackage{tabularx}
\usepackage{makecell}
\usepackage{pifont}                                                     

\usepackage{placeins}

\usepackage{amsfonts}
\usepackage{graphicx}
\usepackage{subfig}
\usepackage{rotating}
\graphicspath{{figures/}}
\usepackage{multirow}
\usepackage{hhline}
\usepackage{amsmath}
\usepackage{adjustbox}
\usepackage{caption}
\usepackage{fancyhdr}
\usepackage{lipsum}  
\usepackage{algorithm}
\usepackage[noEnd=false,indLines=false]{algpseudocodex}

\newcommand{\Input}{\item[\textbf{Input:}]}

\renewcommand{\Return}{\State \textbf{return~}}
\usepackage{booktabs} 
\usepackage{amsmath}

\usepackage{calrsfs}
\DeclareMathAlphabet{\pazocal}{OMS}{zplm}{m}{n}

\newcommand{\RNum}[1]{\lowercase\expandafter{\romannumeral #1\relax}}
\usepackage[font=small]{caption}

\title
{\LARGE \bf
Collapse and Collision Aware Grasping for Cluttered Shelf Picking}

\author{Abhinav Pathak$^{1,2,\dagger}$ and Rajkumar Muthusamy$^{1,\dagger,\ast}$ \thanks{$^{\dagger}$Equal contribution. $^{\ast}$Corresponding Author.} 
\thanks{$^{1}$Robotics Lab, Dubai Future Labs, Dubai, UAE. Email: \tt\small{rajkumar.muthusamy@dubaifuture.gov.ae}} 
\thanks{$^{2}$Birla Institute of Technology and Science, Pilani, Dubai Campus, UAE} 
}

\begin{document}
\maketitle
\thispagestyle{empty}
\pagestyle{empty}

\begin{abstract}

Efficient and safe retrieval of stacked objects in warehouse environments is a significant challenge due to complex spatial dependencies and structural inter-dependencies. Traditional vision-based methods excel at object localization but often lack the physical reasoning required to predict the consequences of extraction, leading to unintended collisions and collapses. This paper proposes a collapse and collision aware grasp planner that integrates dynamic physics simulations for robotic decision-making. Using a single image and depth map, an approximate 3D representation of the scene is reconstructed in a simulation environment, enabling the robot to evaluate different retrieval strategies before execution. Two approaches 1) heuristic-based and 2) physics-based are proposed for both single-box extraction and shelf clearance tasks. Extensive real-world experiments on structured and unstructured box stacks, along with validation using datasets from existing databases, show that our physics-aware method significantly improves efficiency and success rates compared to baseline heuristics.

\end{abstract}

\IEEEpeerreviewmaketitle

\section{Introduction} \label{intro}

\begin{figure}[h!]
    \centering
    \includegraphics[height=0.65\textwidth,width=0.48\textwidth]{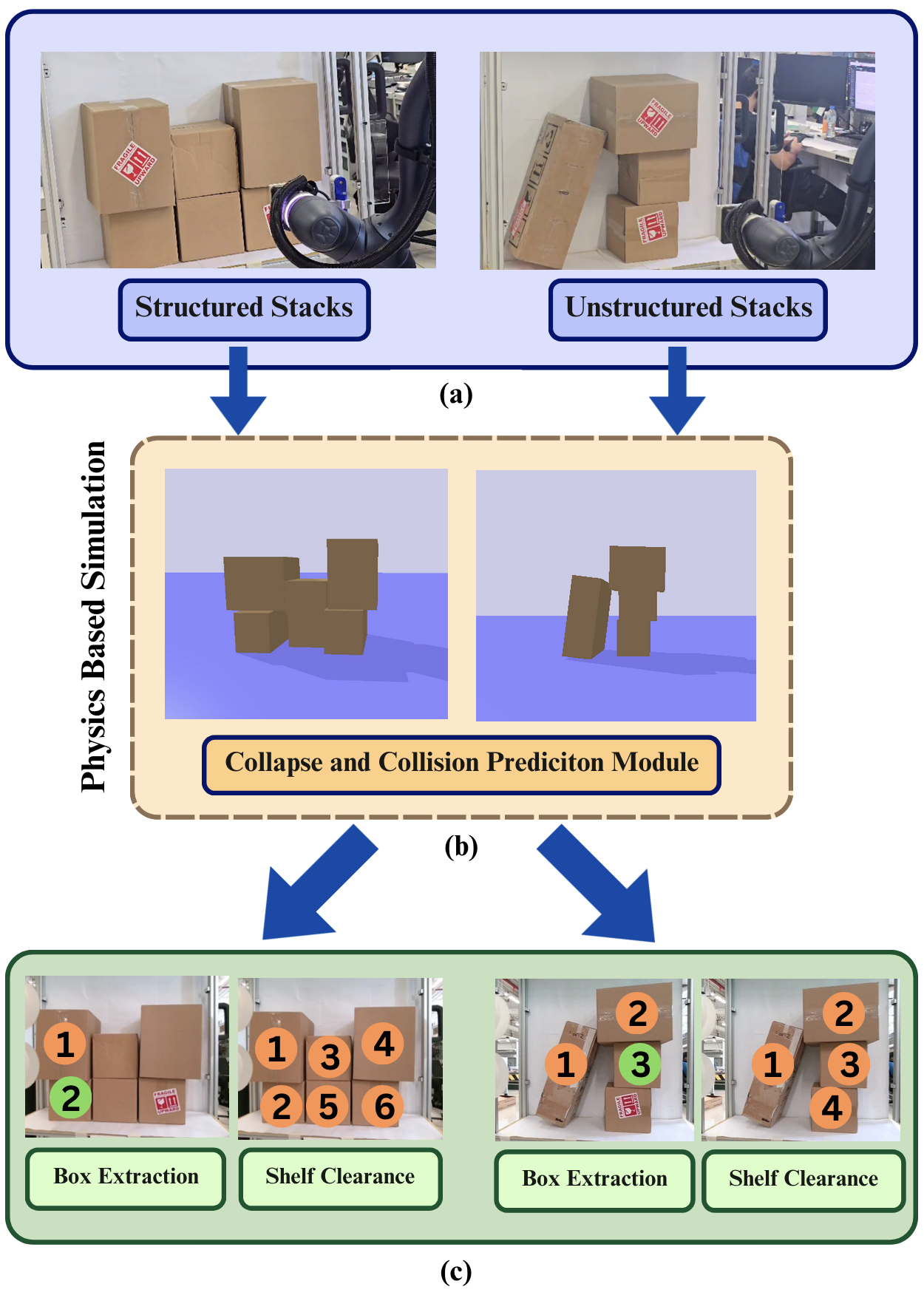}
    \caption{Cluttered shelf object retrieval (a) The autonomous robot observes the stacking complexities (b) using the percieved image  and its features the robot grasp planner conducts physics driven simulation to predict collapses and collison for the retrieval tasks (c) grasp sequence for desired box extraction and  clearing out all boxes from shelf}
    \label{fig:abs}
    \vspace{-1.5em}
\end{figure}

The integration of autonomous robotic systems into warehouse management has led to significant improvements in efficiency, particularly in tasks such as item retrieval, transportation, and inventory organization. A major challenge in such settings is the safe and efficient retrieval of target objects without destabilizing surrounding structures. Unlike traditional pick-and-place operations in robotics, stacked object retrieval involves complex spatial dependencies as seen in Figure \ref{fig:abs}. (a), where the removal of one item can introduce unintended disturbances, leading to collapses or collisions that disrupt warehouse operations. Existing approaches to robotic object retrieval primarily treat the problem as a vision-based task, relying on object detection and segmentation to identify and extract items from storage. 

Although these methods are excellent at identifying object locations, they often overlook the physical effects of removing an object. In particular, they don’t simulate how forces are distributed to keep a stack stable, which can lead to unexpected structural collapses. This lack of predictive reasoning severely limits their applicability in real-world warehouse environments, where errors can cause operational delays, financial losses, and safety hazards.

\begin{figure*}[t!]
    \centering
    \includegraphics[height=0.4\textwidth,width=0.95\textwidth]{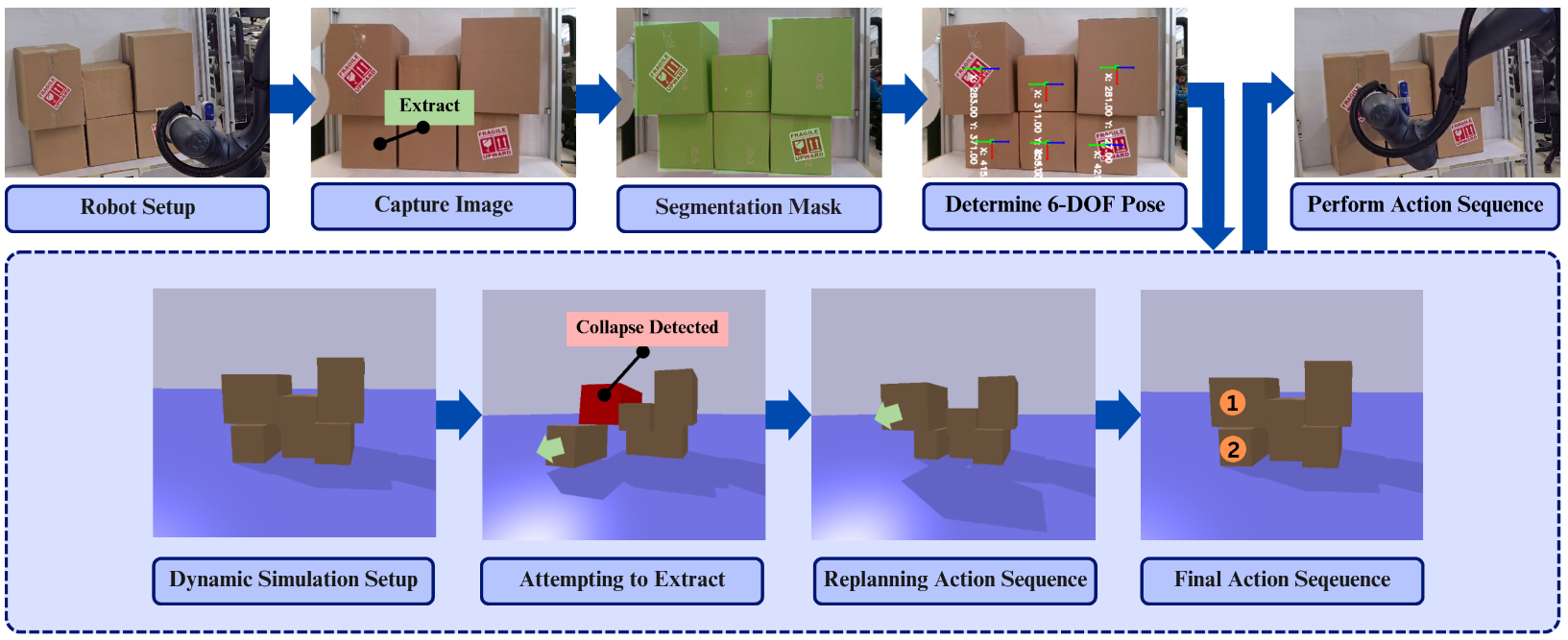}
    \caption{Overview of the proposed grasping pipeline using the physics-aware approach for safe cardboard box extraction}
    \label{fig:graspplanner}
    \vspace{-1.5em}
\end{figure*}

In the context of stacked object retrieval, two primary failure modes must be addressed: 

Collapses: A collapse is a cascading failure where the removal of one object disrupts the weight distribution of the stack, causing multiple objects to fall. This failure mode arises due to a loss of structural support, which can be difficult to predict without physics-aware reasoning. Unlike a direct collision, which involves immediate contact, a collapse can occur after an object is removed due to the gradual redistribution of forces within the stack. 

Collisions: These occur when the robotic manipulator, end-effector, or target object makes unintended contact with adjacent objects during retrieval. Such interactions may cause minor position shifts or major toppling events, leading to downstream failures in structured storage. The problem is exacerbated by occlusions, where objects partially obstruct one another, reducing visibility and increasing the likelihood of contact.

This paper proposes a physics-based approach to model object interactions, enhancing retrieval accuracy and safety. Rather than relying solely on vision-based methods, this system aims to predict the consequences of retrieval actions using a physics-aware real-to-sim approach as in Figure \ref{fig:abs}. (b) and Figure \ref{fig:abs}. (c). With a single RGB and depth image, it can reconstruct an approximate 3D representation of the scene, allowing a robotic arm to evaluate different retrieval strategies before execution. By incorporating segmentation, bounding box estimation, centroid estimation, and depth sensing, it generates a virtual simulation where potential collapses and object movements can be predicted. This allows the robot to develop a form of 'intuition', understanding how the removal of an item will impact the surrounding objects as seen in Figure \ref{fig:graspplanner}.

Our contributions can be summarized as follows:
\vspace{-1.15mm}
\begin{enumerate}
    \item A single shot, collapse-collision aware grasp planner for cluttered shelf picking. The planner conducts physics simulations using the features extracted from the RGB-D image.
    \item Two retrieval approaches, heuristic-based and physics-based, to perform single object extraction and shelf objects clearance.
    \item Extensive real-world experiments to validate and evaluate the planner and approaches under varying stacking complexities (structured and unstructured box stacks). 
    \item Scalability demonstration of the proposed methods on diverse datasets from large-scale databases.
\end{enumerate}

The paper is structured as follows: Section \ref{Related Work} reviews related work in object retrieval, simulation-based planning, and physics-informed robotics. Section \ref{methodology} details the proposed methodology, including the perception pipeline, simulation techniques, and decision-making framework. Experimental results are presented in Section \ref{exp_result}, comparing the performance of the proposed approach in both simulation and real-world scenarios. Finally, Section \ref{conclusions} provides conclusions, discusses limitations, and outlines potential future research directions.

\section{Related Work} \label{Related Work}

Recent advances in robotic manipulation leverage vision, learning, and physics to improve object extraction from clutter. Vision-based methods, like those by Chen et al. \cite{machines11020275} and Mahler et al. \cite{mahler2017dexnet20deeplearning}, use deep neural networks and synthetic data for grasp planning but often neglect structural interactions among objects. Bejjani et al. \cite{bejjani2021occlusionawaresearchobjectretrieval} proposed search strategies for occlusion by predicting object locations, while their learning frameworks \cite{bejjani2019learningphysicsbasedmanipulationclutter} optimize actions in simulation. However, these methods struggled with real-world dynamics. Additionally, Bohg et al. \cite{graspplanning} reviewed various grasp planners for known, familiar and unknown objects. Our collision and collapse predicting method can leverage already existing grasp planners. In this paper we also propose our own simplfied grasp planner. 

Physics-informed approaches incorporate physical reasoning to improve safety and stability. Motoda et al. \cite{9551507} introduced a bi-manual planner that uses physics simulation to predict collapse, a concept extended in \cite{motoda2023multistep} by modeling support relations for safe, multi-step extractions. Meanwhile, real-to-sim frameworks like Zook et al. \cite{zook2024grsgeneratingroboticsimulation} reconstruct 3D scenes from RGB-D data to generate simulation tasks, though they focus more on task synthesis than on ensuring stability.

Other works integrate physics into learning pipelines. Physics-informed neural networks \cite{RAISSI2019686, ni2023progressivelearningphysicsinformedneural} embed physical laws into their models but typically require extensive training and lack real-time adaptability. In contrast, Marchionna et al. \cite{Marchionna_2023} combine instance segmentation with force sensing for precise block extraction in Jenga, while Banerjee et al. \cite{banerjee2024physicsinformedcomputervisionreview} review the potential of physics-aware vision models.

Recent studies have explored hybrid approaches combining vision, learning, and physics, integrating deep neural networks with physical simulation to capture visual cues and structural dynamics. However, these frameworks struggle with real-world adaptability and high computational costs.

Our work bridges these approaches by reconstructing an approximate 3D scene using a real-to-sim pipeline and employing real-time simulation to evaluate extraction strategies. This unified pipeline avoids reliance on pre-trained object models or synthetic datasets \cite{robotics11050104, motoda2023multistep} and generalizes across both structured and unstructured warehouse environments.


\section{Methodology} \label{methodology}

This section presents a physics-aware grasp planner—an end-to-end pipeline that processes visual inputs to generate an executable grasp pose sequence. Additionally, two stacked object retrieval approaches are presented: physics-aware and heuristic-based. These approaches are applied to two key tasks: (1) single-box extraction and (2) shelf clearance. 
The physics-aware approach utilizes simulation to predict potential collapses and collisions, while the heuristic-based approach serves as a baseline for performance evaluation. The following sections provide a detailed breakdown of each component within the proposed planner.

\subsection{Real-to-Sim Pipeline} 

\begin{figure}[h]
    \centering
    \includegraphics[height=0.1\textwidth,width=0.49\textwidth]{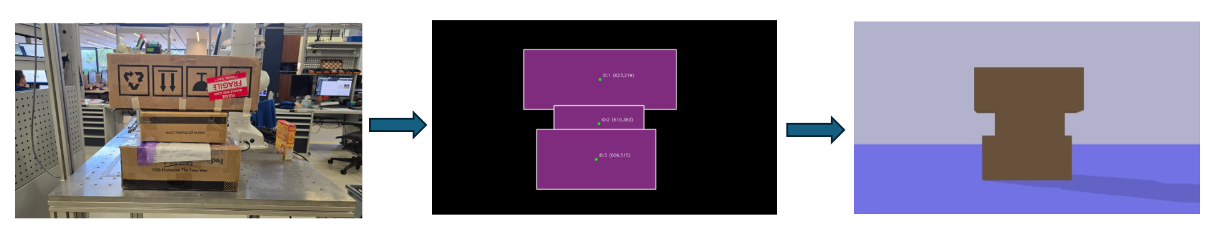}
    \includegraphics[height=0.1\textwidth,width=0.5\textwidth]{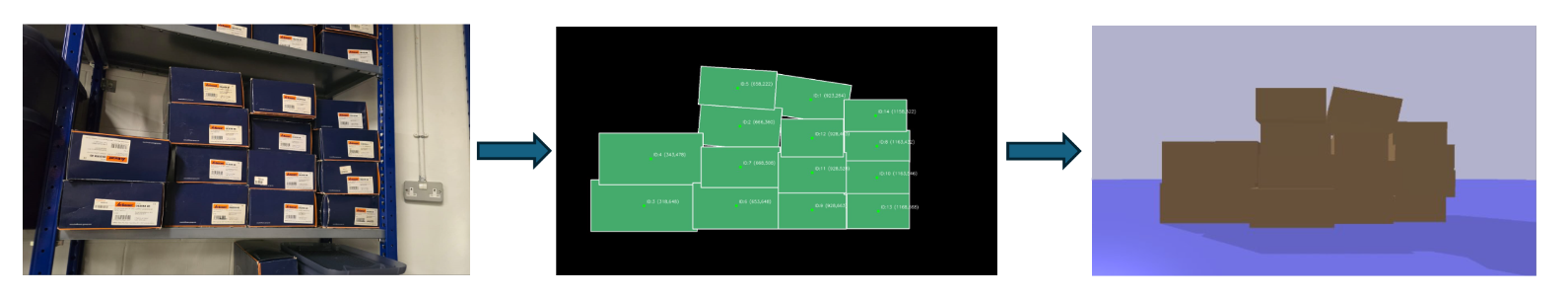}
    \caption{Illustration showcasing how an input image is converted to a simulation scene}
    \label{fig:simtoreal}
\end{figure}

First, a single RGB-D image is captured using a vision sensor, followed by object segmentation. From the resulting mask, features like minimum oriented bounding boxes and centroids are computed, providing estimates of each box’s dimensions, location, and orientation. Finally, integrating the segmentation mask with the depth map, the distance of each box from the camera is also determined. This enables the reconstruction of the observed scene to a 3D simulation space as seen in Figure \ref{fig:simtoreal}. This forms the basis of the physics-aware approach. By simulating the removal of a specific box, the system can predict its impact on the overall stability of the stack. Details of the real-world implementation are provided in the Section \ref{exp_result}.

The output of the perception pipeline reconstructs the approximate positions and orientations of the detected cardboard boxes within the simulation environment. Because the approach relies on a single RGBD image, it is not possible to accurately estimate the depth dimension of each box. However, precise depth measurements are not critical for this approach, which is to capture the relative spatial relationships between boxes in the stack rather than to model them with exact dimensions. To accommodate this uncertainty, the depth of each box is randomized, with the maximum depth constrained by the depth of the shelf. Moreover, the simulation is run ten times, each iteration randomizing the depth values to reflect the natural variability encountered in real-world conditions. This method ensures that the simulation robustly represents the interactions among boxes, enabling reliable predictions of how the removal of any given box will affect the overall stability of the stack.

\subsection{Simulation Properties and Considerations}\label{ch0-Planning-cp0}

PyBullet is used as the physics engine due to its efficiency and ease of integration. Gravity is set to 9.81 m/s², with box density defined as 1 kg/m³ and mass assumed to be uniformly distributed for consistent collision responses. The surface friction coefficient is set to 0.75, while spinning friction is 0.01, ensuring realistic interactions. Friction along the surface is assumed to be constant across all boxes to standardize contact behavior. These parameters enable more realistic simulation of stacked cardboard boxes.

\subsection{Collapse Detection and Simulating Box Removal}\label{ch0-Planning-cp0}

\begin{figure}[h]
    \centering
    \includegraphics[height=0.1\textwidth,width=0.49\textwidth]{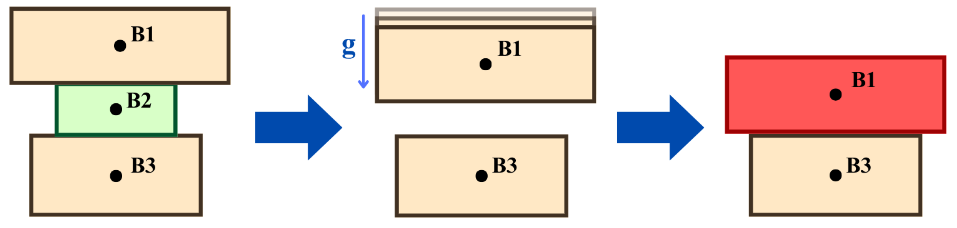}
    \caption{Collapse detection during box removal, highlighting the impact when Box B2 is removed.}
    \label{fig:collapsedetect}
\end{figure}

A box is considered to have collapsed if the simulation detects an unexpected increase in linear or angular velocity beyond a predefined threshold. This enables the simulation to precisely determine the consequences of removing a box in real time. As seen in Figure \ref{fig:collapsedetect}

To further enhance realism, random disturbances and vibrations are introduced during the removal process, accounting for unintended forces that may affect neighboring boxes. Additionally, boxes are removed from the stack by exerting a forward pull on them from their centroids. This is done to mimic how a suction gripper would pull these boxes out. This enables more accurate modeling of how drag and friction affect the stack during the extraction process.

\subsection{Single-Box Removal Using the Physics-Aware Approach}\label{ch0-Planning-cp0}

In this task the goal is to extract a defined target box without causing any unintended collapses or collisions. For this, the physics-aware approach utilizes a backtracking algorithm is used to determine the optimal sequence for removing stacked boxes without causing unintended collapses. The process begins with a user-specified target box, which is the box that needs to be safely extracted. The following explains each step of the algorithm in more detail:


\begin{enumerate}

    \item Initial Removal Attempt:
        The simulation attempts to remove the target box. If this removal does not cause any structural collapse (i.e., no box shifts beyond a predefined movement threshold), the process is complete. However, if a collapse occurs, the simulation resets, and a list of affected (collapsed) boxes is generated.
        
    \item Backtracking to Find a Safe Path:
        From the list of collapsed boxes, a random box (typically the first collapsed box) is selected and removed first. The simulation then checks for any additional collapses. If another collapse occurs, the process repeats: a new list of affected boxes is generated, and another box is removed. This iteration continues until a box is removed that does not cause further collapses.

   \item  Reattempting Target Box Removal:
       Once a stable state is reached (where removing a box does not trigger further collapses), the algorithm retries removing the target box. If removing the target box still leads to a collapse, the backtracking process continues, adjusting the removal order. This recursive approach ensures that boxes are removed in a sequence that prevents structural instability.
    
    \item Final Removal Order:
        The algorithm continues this process until the target box can be safely removed without triggering collapses. The final sequence of removals represents the optimal order for real-world execution. This algorithm is illustrated in figure 5.
\begin{figure}[h]
    \centering
    \includegraphics[height=0.88\textwidth,width=0.6\textwidth]{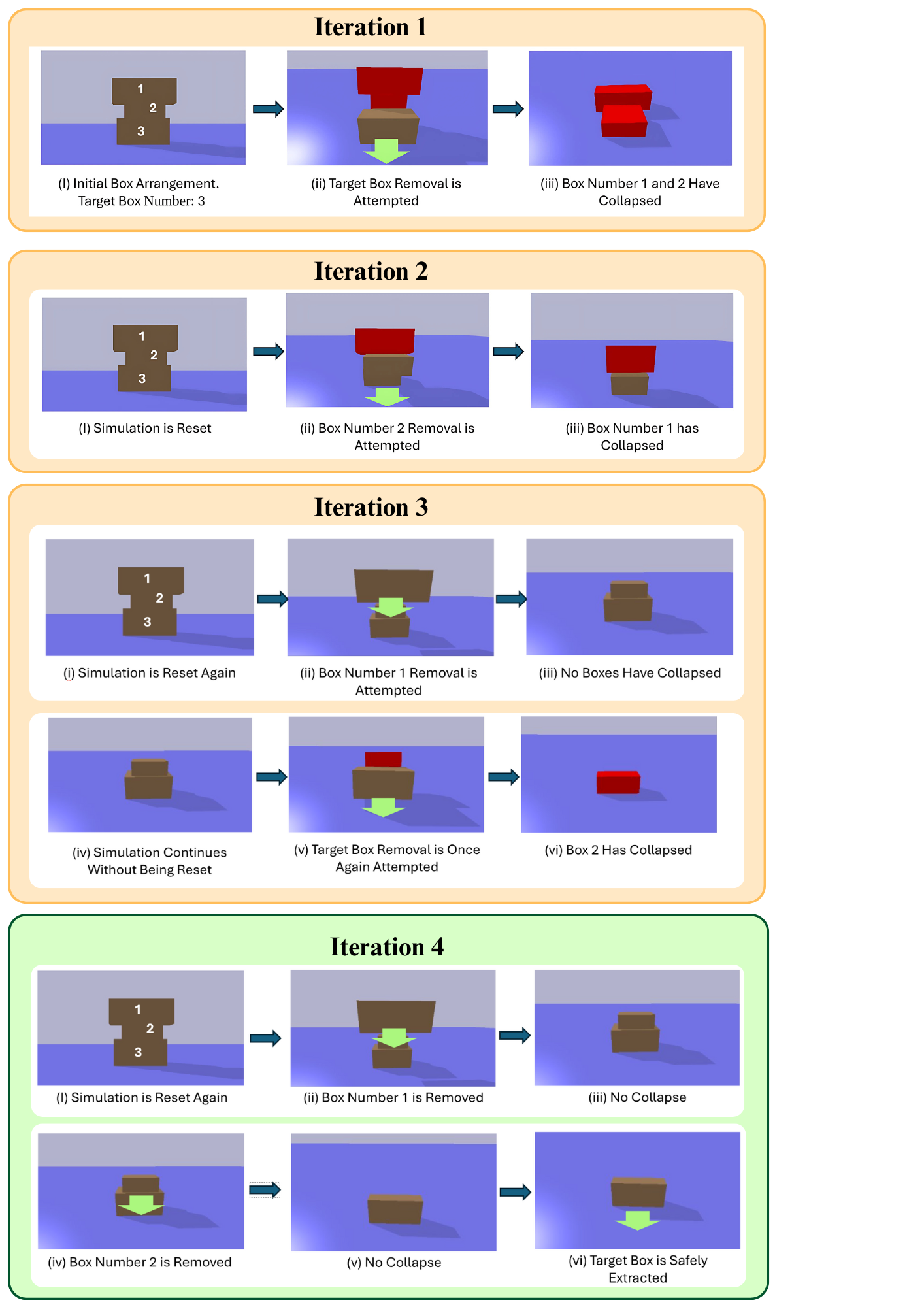}
    \caption{Overview of the single box extraction algorithm}
    \label{fig:simulation_conversion}
\end{figure}

\end{enumerate}
\begin{algorithm}
    \caption{Single-Box Removal Using the Dynamic Simulation}
    \begin{algorithmic}
        \Input RGBD image and target box
        \State $ActionPlan \gets$ Start with an empty array
	    \State $action \gets$ target box
        \While{Solution is not found}
	\State removeBox(action) \Comment{Remove Box as defined by action variable}
         \If{collapse is detected}
	        \State Remove action from ActionPlan
	        \State $action \gets$ First box that collapses
	        \State $resetSimulation()$
         \EndIf
          \If{(collapse is not detected) and (action $\neq$ target box)}
		    \State Add action to ActionPlan
            \State $action \gets$ target box  \Comment {Try to remove target box again}
          \EndIf
        \If {(collapse is not detected) and (action = target box)}
		    \State Solution is Found!
        \EndIf
                \EndWhile
	\Return ActionPlan  
      \end{algorithmic}
\end{algorithm}

\subsection{Shelf Clearance Using the Dynamic Simulation}\label{ch0-Planning-cp0}

This goal of this task is to remove all the boxes in a stack without causing collapse or collisions. Here the proposed approach evaluated each box and removes them one at a time followed by a stability check to detect any unintended changes in the stack. If removal causes instability, such as tilting or shifting, the box is skipped, and the algorithm moves to the next one. If removing a box results in a full stack collapse, the simulation resets to its previous state, and that box is permanently skipped.

If the removal is stable, the box is successfully extracted, and the algorithm continues. This process repeats until all boxes have been evaluated, with only the stable ones being extracted. The algorithm then revisits any skipped boxes and re-evaluates them, following the same checks, until all boxes are removed. Figure 6 and Figure 7 illustrate an overview of the algorithm.

\begin{figure}[h]
    \centering
    \includegraphics[height=0.16\textwidth,width=0.5\textwidth]{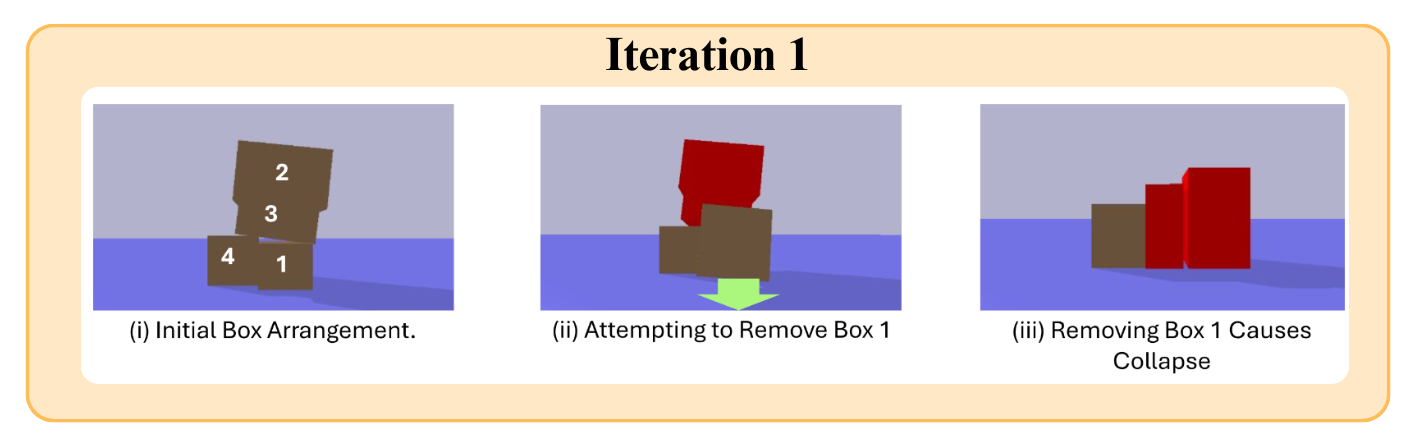}
    \caption{Illustration showing first iteration of the shelf clearing algorithm}
    \label{fig:simulation_conversion}
\end{figure}

\begin{figure}
    \centering
    \includegraphics[height=0.4\textwidth,width=0.5\textwidth]{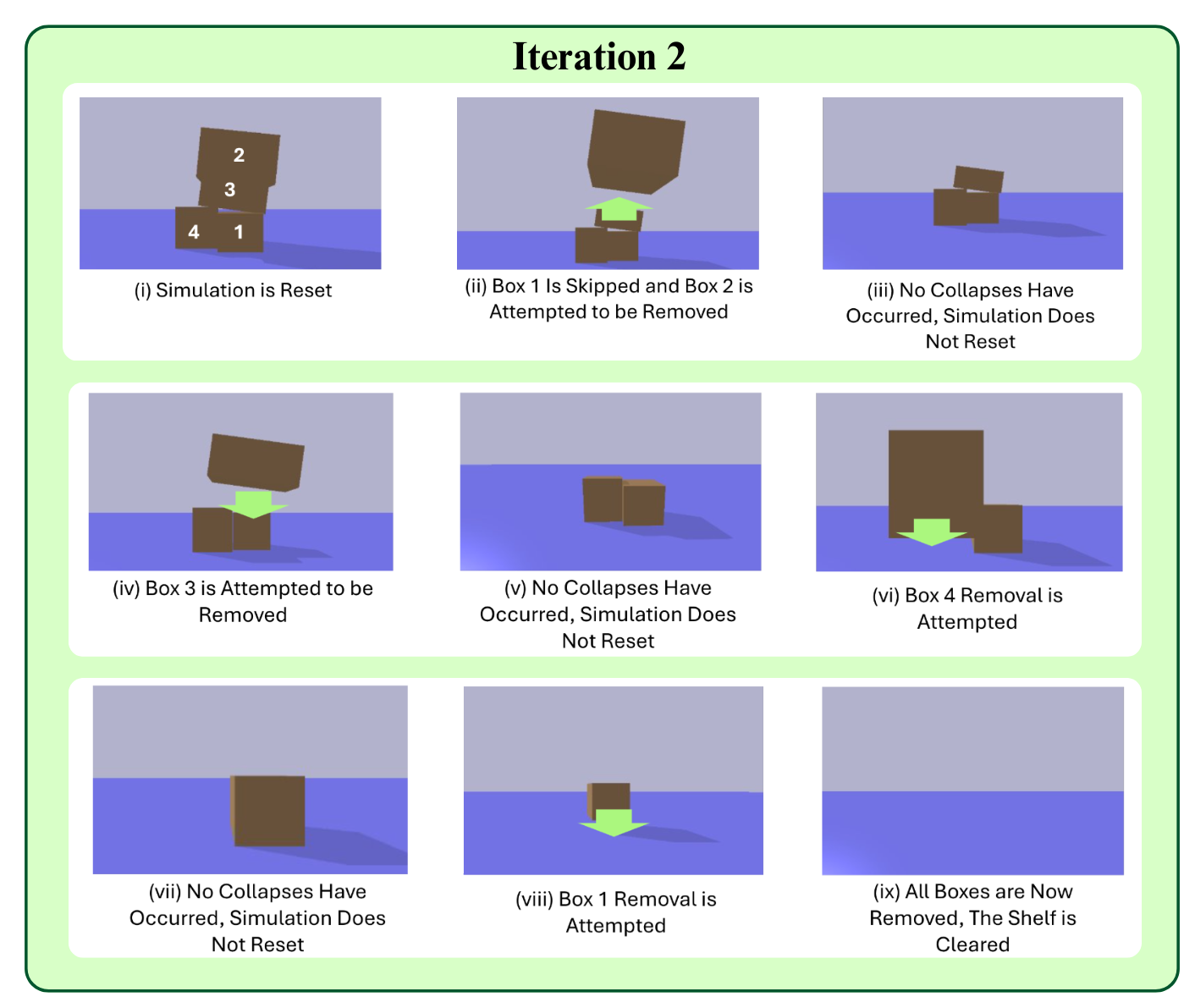}
    \caption{Illustration showing second iteration of the shelf clearing algorithm}
    \label{fig:simulation_conversion}
    \vspace{-1.5em} 
\end{figure}

 \begin{algorithm}
    \captionsetup{skip=5pt}
    \caption{Shelf Clearance Using the Dynamic Simulation}
    \begin{algorithmic}
        \Input RGBD image
        \State $DetectedBoxes \gets$ List of tags of the boxes as detected by the vision pipeline
        \State $ActionPlan \gets$ Start with an empty array
	    \State $action \gets$ first item in DetectedBoxes list
 	    \State $SkippedBoxes \gets$ Start with an empty array 
        \While{Solution is not found}
        \If{action is not in SkippedBoxes}
        \State removeBox(action) \Comment{Remove Box as defined by action variable}
        \EndIf
	
         \If{collapse is detected}
	        \State Remove action from ActionPlan
	        \State Add action to SkippedBoxes list
	        \State $action \gets$ Next box in the DetectedBoxes list
	        \State $resetSimulation()$ \Comment{Resets Simulation}
         \EndIf
          \If{(collapse is not detected) and (action $\neq$ Last Box in DetectedBoxes List )}
		    \State Add action to ActionPlan
            \State $action \gets$ Next box in the DetectedBoxes list
          \EndIf
        \If {(collapse is not detected) and (action = Last Box in DetectedBoxes List)} 
        \If{(SkippedBoxes list is not Empty)}
         \State $SkippedBoxes \gets$ Empty Array
		\State $resetSimulation()$
        \EndIf
        \If{SkippedBoxes list is Empty}
        \State Solution is Found! \Comment{Stop the while loop}
        \EndIf
		   
        \EndIf
                \EndWhile
	\Return ActionPlan  
      \end{algorithmic}
    \end{algorithm}

\FloatBarrier
\subsection{Heuristic-Based Sequence Planning}\label{ch0-Planning-cp0}
This method employs a simple heuristic that considers only the vertical positioning of boxes relative to the ground. It sorts all detected boxes based on their y-axis coordinates, prioritizing those positioned highest in the stack. The system then sequentially removes boxes until it reaches the target. This approach is specifically designed to serve as a baseline to compare the performance of the proposed physics-aware approach and evaluate its effectiveness in more complex scenarios. 

\begin{algorithm}
\caption{Single Box Extraction Using Heuristics}
\begin{algorithmic}
    \State \textbf{Input:} RGB Image and Target Box
    \State $positions \gets$ Coordinates of the centroids of all detected boxes
    \State $SortedPositions \gets$ sorted positions \Comment{Sort positions from highest to lowest}
    \State $ActionPlan \gets$ Start with an empty array
    \For{\textbf{each} $item$ in $SortedPositions$}
        \If{$action \neq target$}
            \State Add $action$ to $ActionPlan$
        \EndIf
        \If{$action = target$}
            \break
        \EndIf
    \EndFor
    \State \Return $ActionPlan$
\end{algorithmic}
\end{algorithm}

\begin{figure}[h]
    \centering
    \includegraphics[height=0.18\textwidth,width=0.45\textwidth]{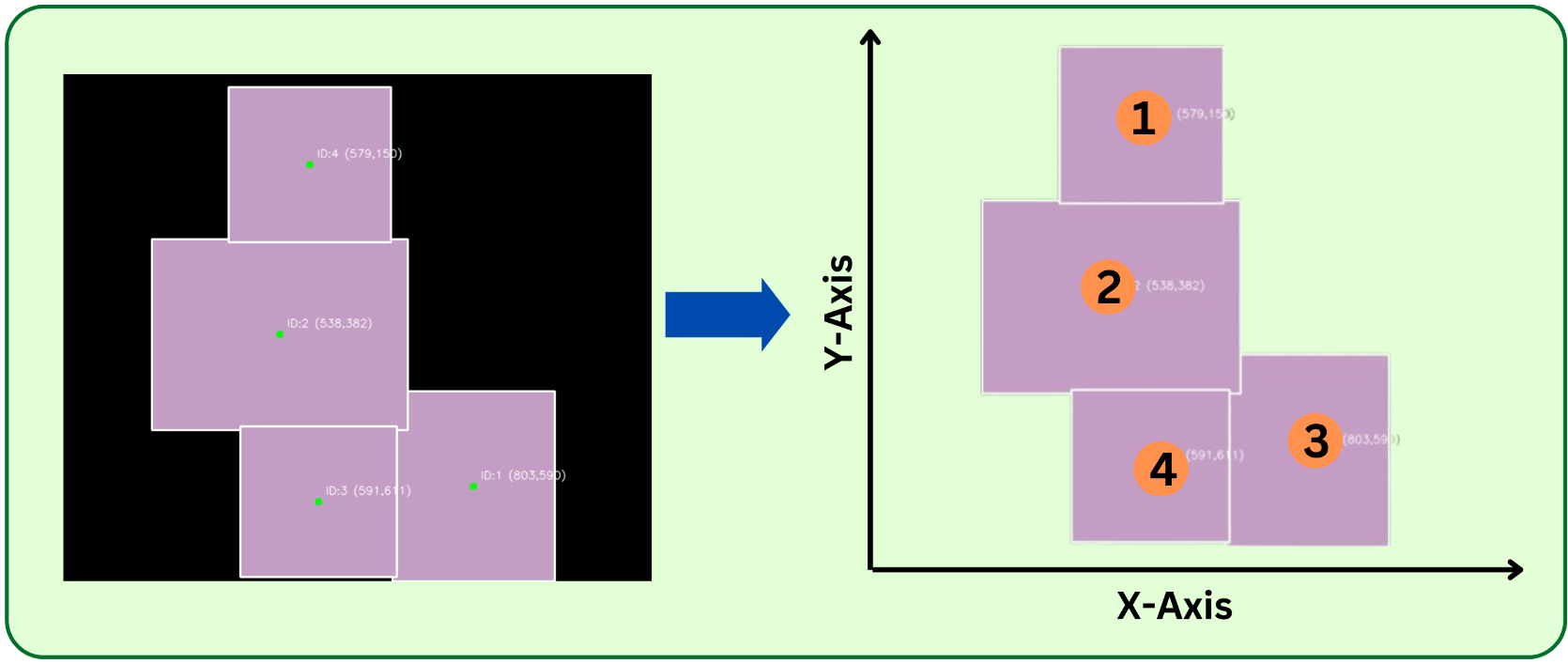}
    \caption{Illustration showing the action plan generated by the base heuristics approach for a shelf clearing task}
    \label{fig:base}
    \vspace{-1.5em}
\end{figure}

For shelf clearance, the algorithm keeps removing boxes until all the boxes are removed from the shelf. As seen in Figure \ref{fig:base}

\section{Experiments and Results} \label{exp_result}


\subsection{Experimental Setup}\label{ch0-Planning-cp0}

\begin{figure}[h]
    \centering
    \includegraphics[height=0.18\textwidth,width=0.45\textwidth]{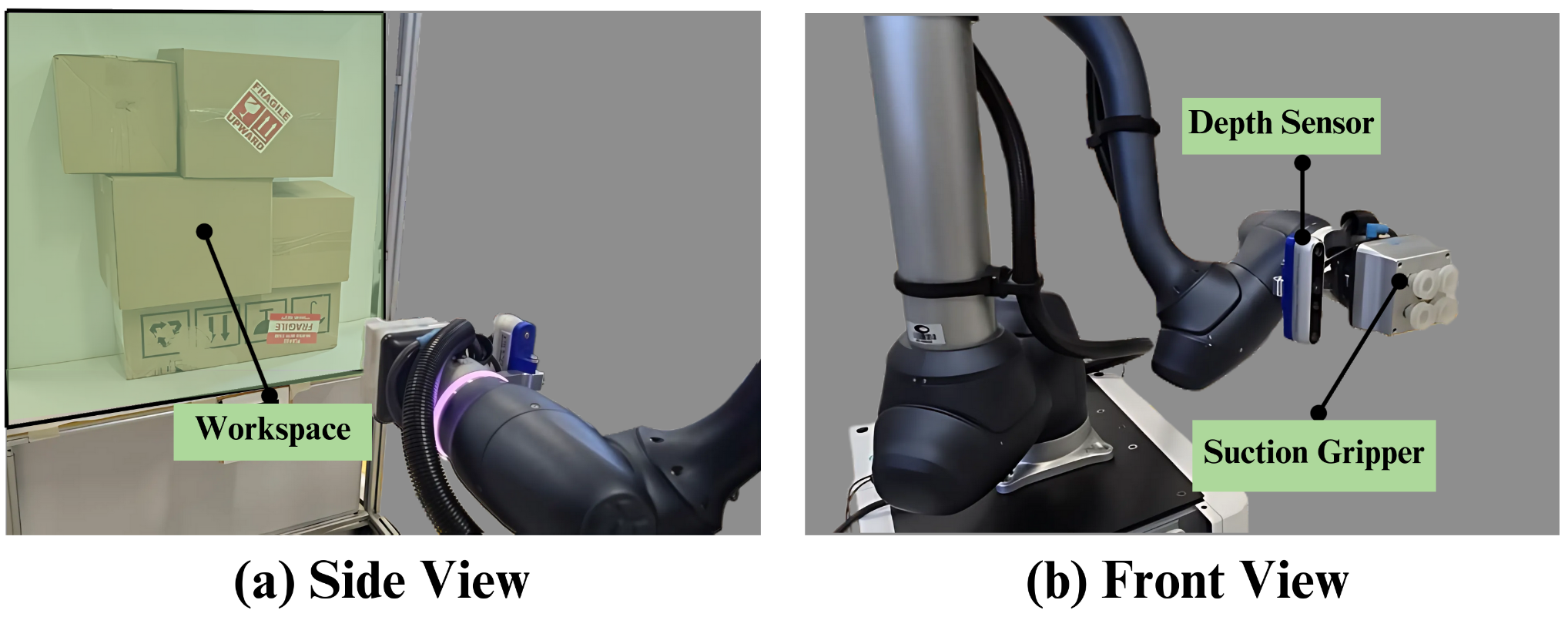}
    \caption{Experimental Setup}
    \label{fig:setup}
    \vspace{-1.5em}
\end{figure}

The experimental setup (shown in Figure \ref{fig:setup} features a Doosan H2515 6-degree-of-freedom (6DOF) robotic arm, equipped with a suction gripper for precise box handling, and an Intel RealSense D455 depth camera for real-time spatial perception. A 100 cm × 30 cm × 160 cm shelf, positioned 104 cm from the robot, served as the designated area for stacking boxes and was used for both single-box extraction and shelf clearance experiments. To demonstrate the robustness of the proposed approach, mixed-sized cardboard boxes were utilized to create the stacking scenes. Specifically, three types of cardboard boxes with varying dimensions—23 cm × 31 cm × 25 cm, 20 cm × 20 cm × 20 cm, and 50 cm × 17 cm × 17 cm, were selected. These different box sizes introduce variability in the stacking environment.

\subsection{Considered Stacking Complexities}\label{stacking}

\begin{figure}[h]
    \centering
    \includegraphics[height=0.15\textwidth,width=0.47\textwidth]{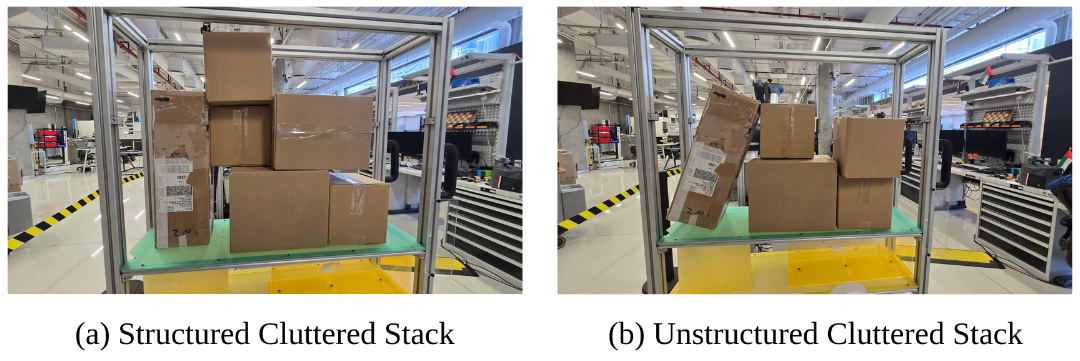}
    \caption{Different Stacking Complexities Considered }
    \vspace{-1.5em}
    \label{fig:complex}
\end{figure}

To better define the problem space, this study categorizes stacked structures into two distinct types:

\begin{enumerate}
    \item \textbf{Structured Cluttered Stacks} – While these stacks maintain a general order, they introduce variations in depth, meaning some boxes extend further out than others. This uneven layering creates partial obstructions, reducing visibility and making removal more challenging. This is shown in Figure \ref{fig:complex} (a) 
    \item\textbf{Unstructured Cluttered Stacks} - These are the more complex, consisting of randomly placed boxes with varying orientations, rotations, and significant depth variations. The unpredictability of these stacks increases the risk of collapse when removing boxes. This is shown in Figure \ref{fig:complex} (b) 
\end{enumerate}

\begin{figure*}[b]
    \centering
    \includegraphics[width=0.8\textwidth,height=0.36\textwidth]{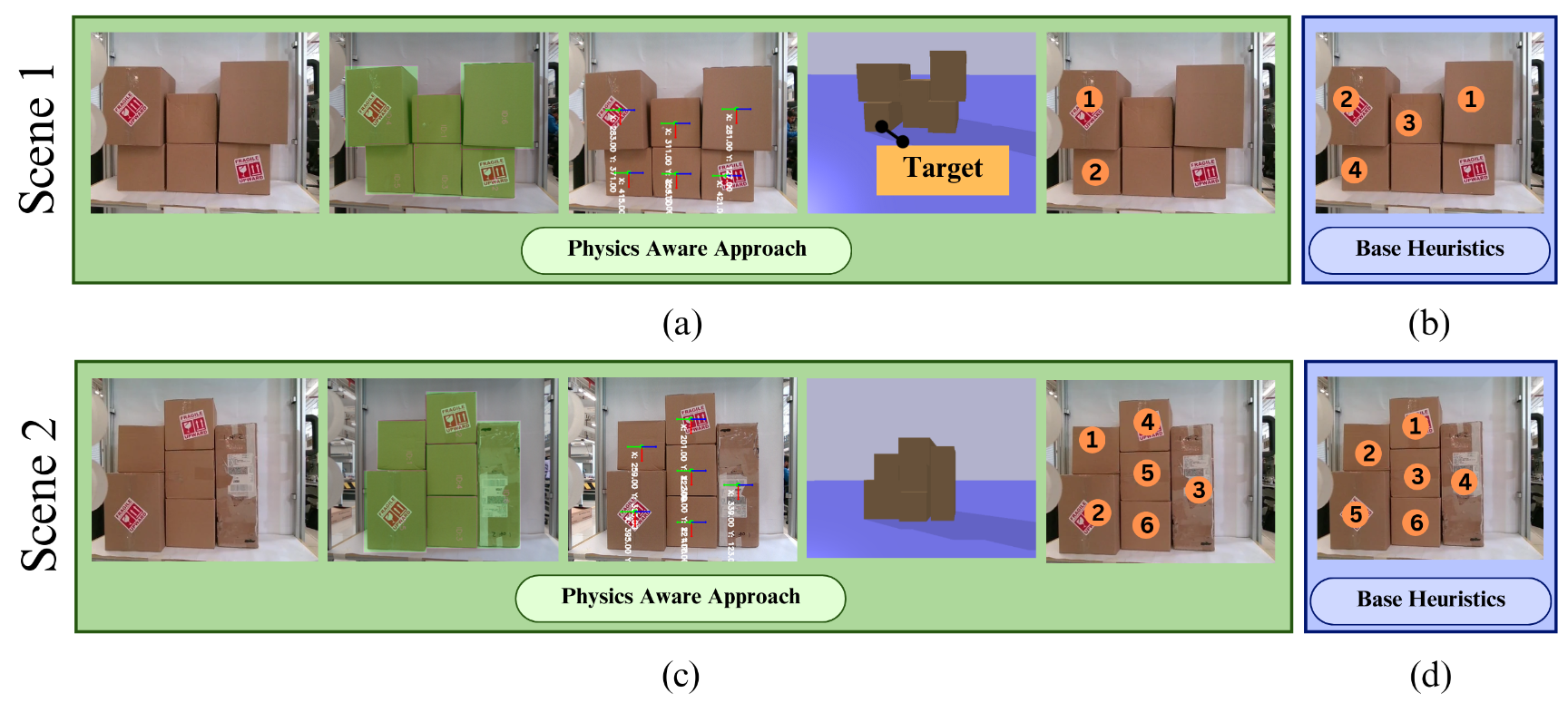}
    \caption{ Sequence planning for structured stack experiments: (a) illustrates the physics-based approach for the extraction task, (b) presents action sequences predicted by the baseline heuristics approach, (c) depicts the physics-aware method’s action sequence for shelf clearance, and (d) shows the output predicted by the baseline heuristics approach }
    \vspace{-1.5em}
    \label{fig:scene12}
\end{figure*}

\subsection{Training Settings}\label{ch0-Planning-cp0}
As explained in the previous section, the boxes are first perceived by the camera and then placed into the simulation space. To achieve this, a single RGBD image is captured using an Intel RealSense D455 depth camera. Object segmentation is performed using a custom-trained YOLOv11n-seg model \cite{yolo11_ultralytics} trained on the Online Stacked Cardboard Boxes Dataset \cite{yang2021scdstackedcartondataset}, comprising 8,401 images. The dataset was split into a training set (80\%), validation set (10\%), and test set (10\%) to ensure robust model evaluation. Given the dataset's limited variety in box orientations, data augmentation was employed to improve the model's ability to handle angled boxes. Specifically, random rotations ranging from -90 to 90 degrees were applied to the training images to simulate different box angles and enhance generalization. The model was trained on a system equipped with an RTX 3070 Ti (8GB VRAM), paired with an Intel i7-11700K processor and 64GB of RAM. During training, the model achieved a Mean Average Precision (mAP) of 0.87 on the test set, with validation performance consistently matching the test set results, indicating the model’s generalization capability.




\begin{figure*}[t!]
    \centering
    \includegraphics[height=0.36\textwidth,width=0.8\textwidth]{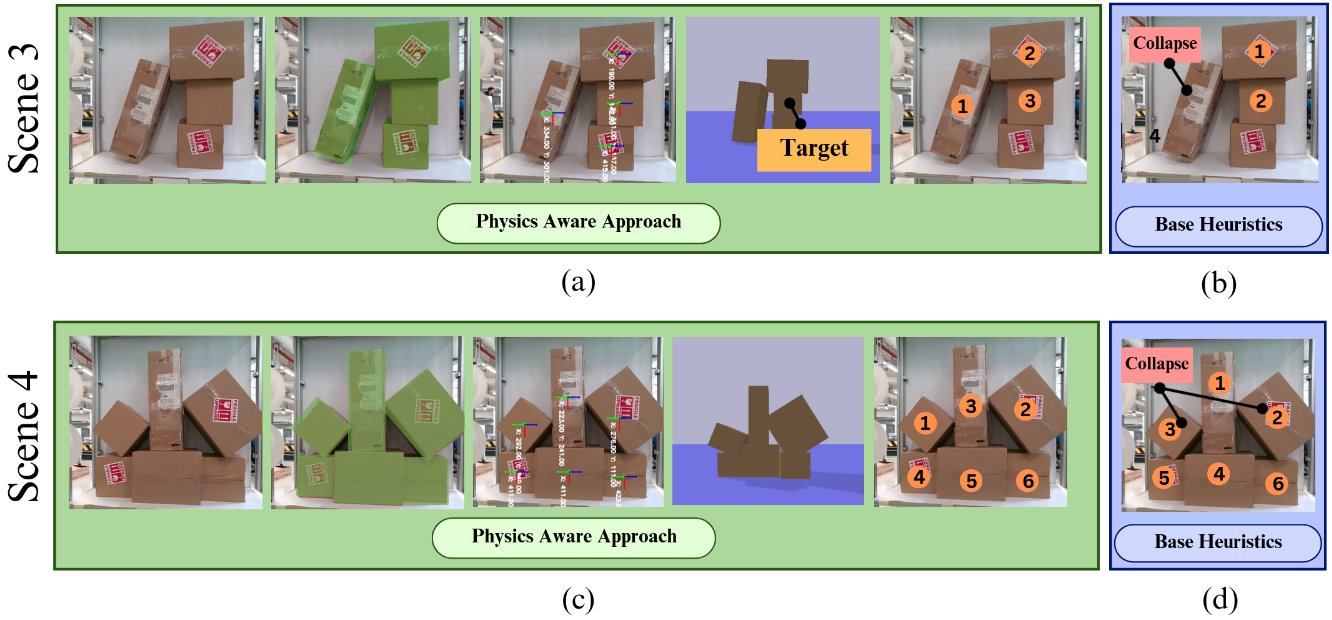}
    \caption{Sequence planning for unstructured stack experiments: (a) demonstrates the physics-based approach for the extraction task, (b) showcases action sequences predicted by the baseline heuristics approach, (c) illustrates the physics-aware method’s action sequence for shelf clearance, and (d) presents the output predicted by the baseline heuristics approach}
    \label{fig:scene34}
    \vspace{-1.5em}
\end{figure*}

\begin{table}[!b]
\vspace{-1.5em}
  \centering
  \caption{Experimental Results for Single Box Extraction Tasks}
  \label{tab:single_box_extraction}
  \renewcommand{\arraystretch}{1.2} 
  \begin{tabularx}{\linewidth}{|c|>{\centering\arraybackslash}m{2cm}|c|c|>{\centering\arraybackslash}m{1.2cm}|}
    \hline
    \textbf{Condition} & \textbf{Approach} & \textbf{Result} & \textbf{Time (s)} & \textbf{Boxes Removed} \\
    \hline
    \multirow{2}{*}{Structured} 
      & Physics-Aware   & \ding{51} & 43 & 2 \\ \cline{2-5}
      & Base Heuristics & \ding{51} & 88 & 4 \\ 
    \hline
    \multirow{2}{*}{Unstructured} 
      & Physics-Aware   & \ding{51} & 75 & 3 \\ \cline{2-5}
      & Base Heuristics & Fail & 38 & 2 \\ 
    \hline
  \end{tabularx}
\end{table}

\begin{table}[!b]
\vspace{-1.2em}
  \centering
  \caption{Experimental Results for Shelf Clearance Tasks}
  \label{tab:shelf_clearance}
  \begin{tabular}{|c|c|c|c|}
    \hline
    \textbf{Condition} & \textbf{Approach} & \textbf{Result} & \textbf{Time (s)} \\
    \hline
    \multirow{2}{*}{Structured} 
      & Physics-Aware   & \ding{51} & 116 \\ \cline{2-4}
      & Base Heuristics & \ding{51} & 98 \\ 
    \hline
    \multirow{2}{*}{Unstructured} 
      & Physics-Aware   & \ding{51} & 110 \\ \cline{2-4}
      & Base Heuristics &  Fail   & 25 \\ 
    \hline
  \end{tabular}
\end{table}

\subsection{Results}\label{ch0-Planning-cp0}
Experiments were conducted testing the efficacy of the proposed approach on two tasks: (1) single box extraction and (2) shelf clearance. In the single box extraction task, the objective was to remove a specific target box without causing any collapse or collision. In the shelf clearance task, the goal is to efficiently extract all boxes from a shelf while maintaining structural integrity. The study conducted eight experiments across two tasks and two stacking conditions using both physics-aware and base heuristics approaches as follows:

\begin{enumerate}
    \item Scene 1: Single box extraction in structured stacks using physics-aware and base-heuristics approach as seen in Figure \ref{fig:scene12} (a) and (b)
    \item Scene 2: Shelf clearance in structured stacks using physics-aware and base-heuristics approach as seen in Figure \ref{fig:scene12} (c) and (d). 
    \item Scene 3: Single box extraction in unstructured stacks using physics-aware and base-heuristics approach as seen in Figure \ref{fig:scene34} (a) and (b). 
    \item Scene 4: Shelf clearance in unstructured stacks using physics-aware and base-heuristics approach as seen in Figure \ref{fig:scene34} (c) and (d). 
\end{enumerate}



The results demonstrate that in structured stacks, both methods extract the target box successfully, although the physics-aware approach achieves higher efficiency by removing fewer extraneous boxes. In structured shelf clearance, both methods clear the shelf without incident. However, in unstructured stacks, the base heuristics approach frequently fails to extract the target box and clear the shelf, whereas the physics-aware approach consistently achieves successful outcomes. This is because the physics-aware approach allows for a more accurate modeling of the relationships between boxes in the stack. As a result, the algorithm can detect potential collapses even when the affected boxes are not directly stacked on top of one another. The results from these experiments are shown in Table \ref{tab:single_box_extraction} and Table \ref{tab:shelf_clearance}. 

\subsection{Scalability} \label{data-scaling}
\begin{figure*}[!bt]
    \centering
    \includegraphics[height=0.22\textwidth,width=0.9\textwidth]{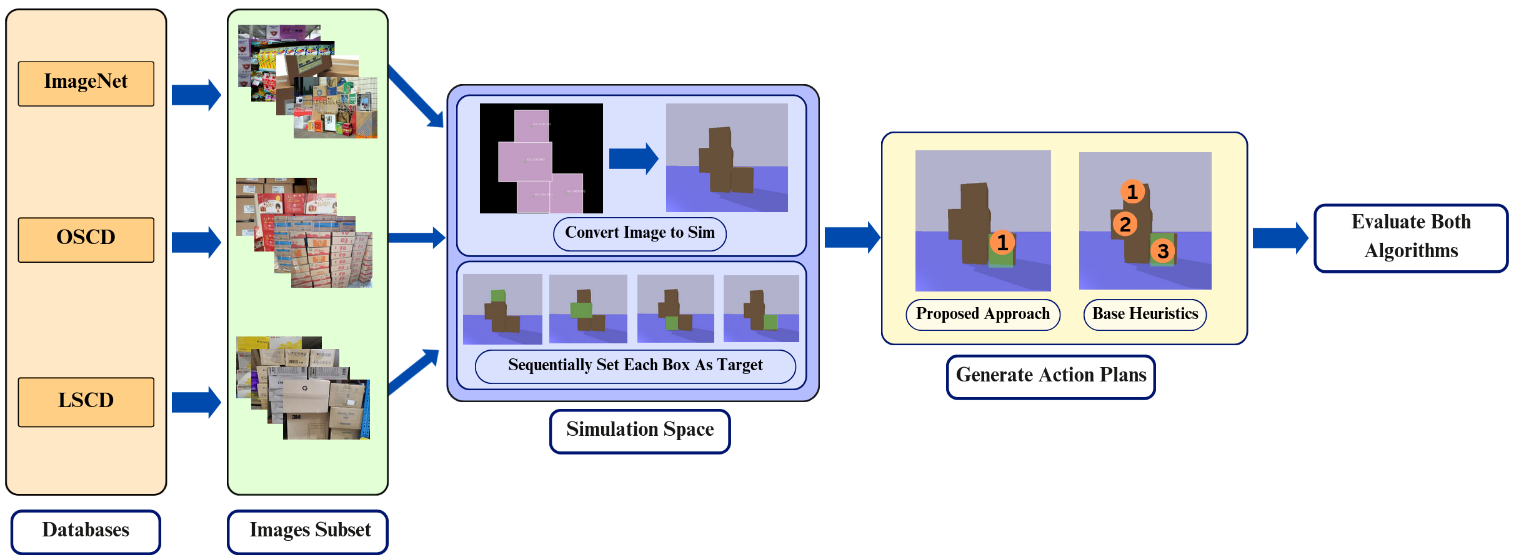}
    \caption{Overview of the scalability experiments }
    \label{fig:scale}

\end{figure*}

Following the real-world experimental results, evaluation on three diverse datasets was conducted to assess the scalability and performance of the proposed approach relative to base heuristics. The evaluation used 500 images from ImageNet \cite{imagenet}, 1,000 images from the Online Stacked Carton Dataset (OSCD) \cite{yang2021scdstackedcartondataset} (which consisted solely of images from the test and validation sets) and 1,000 images from the Live Stacked Carton Dataset (LSCD)\cite{yang2021scdstackedcartondataset}. For each image, the vision pipeline converted the scene into a simulation environment. In each simulation, every box was sequentially designated as the target. For each target, both the physics-aware approach and base heuristics generated action plans, which were executed in the simulation to evaluate efficiency and detect any unintended collisions or collapses .Efficiency was calculated using equation \ref{eq:efficiency_improvement}.  The results are collated in Table \ref{tab:scaling_results}, with an overview shown in Figure \ref{fig:scale}.\begin{equation}
\text{Efficiency Improvement (\%)} = \left(\frac{T_{bh} - T_{pa}}{T_{pa}}\right) \times 100
\label{eq:efficiency_improvement}
\end{equation}
Here, \(T_{bh}\) is the average time taken by the base-heuristics approach and \(T_{pa}\) is the average time taken by the physics-aware approach.

\begin{table*}[ht]
    \caption{Comparison of Base Heuristics and Physics-Aware Approach}
    \label{tab:scaling_results}
    \centering
    \renewcommand{\arraystretch}{1.2} 
    \begin{tabularx}{\textwidth}{|>{\centering\arraybackslash}m{3.3cm}|*{2}{>{\centering\arraybackslash}m{2cm}|}*{4}{>{\centering\arraybackslash}m{1.85cm}|}}
         \hline
         \multirow{2}{*}{\textbf{Data Subset}} & \multicolumn{2}{c|}{\textbf{Base Heuristics (B.H)}} & \multicolumn{4}{c|}{\textbf{Physics-Aware Approach}} \\
         \cline{2-7}
         & \textbf{No. of Boxes Removed (Avg.)} & \textbf{Avg. Execution Time (s)} & \textbf{No. of Boxes Removed (Avg.)} & \textbf{Avg. Execution Time (s)} & \textbf{Success Rate Relative to B.H} & \textbf{Efficiency Relative to B.H} \\
         \hline
         ImageNet \cite{imagenet} (500 Images) & 4.98 & 56.16 & 2.51 & 37.56 & +43\% & +49.52\% \\
         \hline
         OSCD \cite{yang2021scdstackedcartondataset} (1000 Images)    & 5.09 & 64.08 & 2.57 & 32.06 & +53.3\% & +96.76\% \\
         \hline
         LSCD \cite{yang2021scdstackedcartondataset} (1000 Images)   & 4.94 & 62.62 & 2.55 & 31.70 & +61.20\% & +94.56\% \\
         \hline
    \end{tabularx}
\end{table*}

\vspace{-0.5em}
\subsection{Discussions}\label{ch0-Planning-cp0}

The physics-aware approach consistently outperformed base heuristics across environments and datasets. In structured settings, both methods completed tasks, but the physics-aware approach was significantly more efficient, extracting boxes in 43 seconds with minimal disruption (removing only 2 boxes) versus 88 seconds and 4 removals for base heuristics. This efficiency gap widened in unstructured environments, where base heuristics failed, while the physics-aware approach remained reliable.

Scaling across datasets reinforced these trends. As shown in Table~\ref{tab:scaling_results}, the physics-aware method consistently removed fewer boxes, completed tasks nearly twice as fast, and improved success rates by up to 61.2\%, and retrieval efficiency of upto 96.76 \%.

While base heuristics perform in structured environments, they struggle with real-world complexity. By modeling indirect interactions, the physics-aware approach proves more robust and scalable for robotic manipulation. Minor segmentation inaccuracies from single-depth perception can introduce occasional errors, but overall, integrating physical modeling significantly enhances efficiency and reliability.

\section{Conclusions} \label{conclusions}

This paper presented a physics-aware grasp planner for retrieving diversely cluttered stacked boxes from shelves. Real world experiments using a robotic manipulator equipped with a suction gripper compared and validated the practical effectiveness of the proposed physics-aware and heuristics-based approaches for safe shelf picking. Although the current implementation uses suction gripper, the proposed physics-aware approach can also be extended to other grasping pipelines and end-effectors. The experiments show that in structured stacks, the base-heuristics approach is less efficient, and in unstructured conditions, it fails completely. While the physics-aware approach proved more efficient and successful, it required longer computation times than the base-heuristics method.

Additionally, evaluations on large scale existing datasets showed that the physics-aware approach consistently outperforms the base-heuristics approach, achieving up to 61.2\% higher success rates and 96.76\% greater efficiency, particularly in unstructured settings.

A key limitation of the current system lies in the perception pipeline, which relies on a single depth image and can lead to misaligned boxes in simulation. Future work will can include bi-manual manipulation, multi-view sensing, visual servoing, and adaptive planning to enhance robustness and scalability across a broader range of warehouse scenarios.

\section*{Acknowledgments}
This work is an outcome of the BITS(Dubai)-DFL research project. We would like to thank Dr. Tarek Taha and Prof. Kalaichelvi Venkatesan for their valuable advisory support.

\bibliographystyle{IEEEtran}
\bibliography{references}  

\begin{thebibliography}{10}
\providecommand{\url}[1]{#1}
\csname url@samestyle\endcsname
\providecommand{\newblock}{\relax}
\providecommand{\bibinfo}[2]{#2}
\providecommand{\BIBentrySTDinterwordspacing}{\spaceskip=0pt\relax}
\providecommand{\BIBentryALTinterwordstretchfactor}{4}
\providecommand{\BIBentryALTinterwordspacing}{\spaceskip=\fontdimen2\font plus
\BIBentryALTinterwordstretchfactor\fontdimen3\font minus \fontdimen4\font\relax}
\providecommand{\BIBforeignlanguage}[2]{{%
\expandafter\ifx\csname l@#1\endcsname\relax
\typeout{** WARNING: IEEEtran.bst: No hyphenation pattern has been}%
\typeout{** loaded for the language `#1'. Using the pattern for}%
\typeout{** the default language instead.}%
\else
\language=\csname l@#1\endcsname
\fi
#2}}
\providecommand{\BIBdecl}{\relax}
\BIBdecl

\bibitem{machines11020275}
Y.-L. Chen, Y.-R. Cai, and M.-Y. Cheng, ``Vision-based robotic object grasping—a deep reinforcement learning approach,'' \emph{Machines}, vol.~11, no.~2, 2023.

\bibitem{mahler2017dexnet20deeplearning}
J.~Mahler, J.~Liang, S.~Niyaz, M.~Laskey, R.~Doan, X.~Liu, J.~A. Ojea, and K.~Goldberg, ``Dex-net 2.0: Deep learning to plan robust grasps with synthetic point clouds and analytic grasp metrics,'' 2017.

\bibitem{bejjani2021occlusionawaresearchobjectretrieval}
\BIBentryALTinterwordspacing
W.~Bejjani, W.~C. Agboh, M.~R. Dogar, and M.~Leonetti, ``Occlusion-aware search for object retrieval in clutter,'' 2021. [Online]. Available: \url{https://arxiv.org/abs/2011.03334}
\BIBentrySTDinterwordspacing

\bibitem{bejjani2019learningphysicsbasedmanipulationclutter}
W.~Bejjani, M.~R. Dogar, and M.~Leonetti, ``Learning physics-based manipulation in clutter: Combining image-based generalization and look-ahead planning,'' 2019.

\bibitem{graspplanning}
J.~Bohg, A.~Morales, T.~Asfour, and D.~Kragic, ``Data-driven grasp synthesis—a survey,'' \emph{IEEE Transactions on Robotics}, vol.~30, no.~2, pp. 289--309, 2014.

\bibitem{9551507}
T.~Motoda, D.~Petit, W.~Wan, and K.~Harada, ``Bimanual shelf picking planner based on collapse prediction,'' in \emph{2021 IEEE 17th International Conference on Automation Science and Engineering (CASE)}, 2021, pp. 510--515.

\bibitem{motoda2023multistep}
T.~Motoda, D.~Petit, T.~Nishi, K.~Nagata, W.~Wan, and K.~Harada, ``Multi-step object extraction planning from clutter based on support relations,'' \emph{IEEE Access}, vol.~PP, pp. 1--1, 01 2023.

\bibitem{zook2024grsgeneratingroboticsimulation}
A.~Zook, F.-Y. Sun, J.~Spjut, V.~Blukis, S.~Birchfield, and J.~Tremblay, ``Grs: Generating robotic simulation tasks from real-world images,'' 2024.

\bibitem{RAISSI2019686}
M.~Raissi, P.~Perdikaris, and G.~Karniadakis, ``Physics-informed neural networks: A deep learning framework for solving forward and inverse problems involving nonlinear partial differential equations,'' \emph{Journal of Computational Physics}, vol. 378, pp. 686--707, 2019.

\bibitem{ni2023progressivelearningphysicsinformedneural}
R.~Ni and A.~H. Qureshi, ``Progressive learning for physics-informed neural motion planning,'' 2023.

\bibitem{Marchionna_2023}
L.~Marchionna, G.~Pugliese, M.~Martini, S.~Angarano, F.~Salvetti, and M.~Chiaberge, ``Deep instance segmentation and visual servoing to play jenga with a cost-effective robotic system,'' \emph{Sensors}, no.~2, 2023.

\bibitem{banerjee2024physicsinformedcomputervisionreview}
C.~Banerjee, K.~Nguyen, C.~Fookes, and G.~Karniadakis, ``Physics-informed computer vision: A review and perspectives,'' 2024.

\bibitem{robotics11050104}
T.~Motoda, D.~Petit, T.~Nishi, K.~Nagata, W.~Wan, and K.~Harada, ``Shelf replenishment based on object arrangement detection and collapse prediction for bimanual manipulation,'' \emph{Robotics}, vol.~11, no.~5, 2022.

\bibitem{yolo11_ultralytics}
\BIBentryALTinterwordspacing
G.~Jocher and J.~Qiu, ``Ultralytics yolo11,'' 2024. [Online]. Available: \url{https://github.com/ultralytics/ultralytics}
\BIBentrySTDinterwordspacing

\bibitem{yang2021scdstackedcartondataset}
J.~Yang, S.~Wu, L.~Gou, H.~Yu, C.~Lin, J.~Wang, M.~Li, and X.~Li, ``Scd: A stacked carton dataset for detection and segmentation,'' 2021.

\bibitem{imagenet}
J.~Deng, W.~Dong, R.~Socher, L.-J. Li, K.~Li, and L.~Fei-Fei, ``Imagenet: A large-scale hierarchical image database,'' in \emph{2009 IEEE Conference on Computer Vision and Pattern Recognition}, 2009, pp. 248--255.

\end{thebibliography}

\end{document}